\documentclass{article}
\usepackage{PRIMEarxiv}
\pdfoutput=1
\usepackage[utf8]{inputenc} 
\usepackage[T1]{fontenc}    
\usepackage{hyperref}       
\usepackage{url}            
\usepackage{booktabs}       
\usepackage{amsfonts}       
\usepackage{nicefrac}       
\usepackage{microtype}      
\usepackage{lipsum}
\usepackage{fancyhdr}       
\usepackage{graphicx}       
\usepackage[nocompress]{cite}

\graphicspath{{media/}}     

\usepackage[american]{babel}

\usepackage{mathtools} 
\usepackage{tikz} 
\usepackage{times}
\usepackage{soul}
\usepackage{amsmath}
\usepackage{array}
\usepackage{amsthm}
\usepackage{algorithm}
\usepackage{algorithmic}
\usepackage[switch]{lineno}
\usepackage{subcaption}
\usepackage{multirow}
\usepackage{svg}
\usepackage{amssymb}  
\usepackage{amsfonts}
\usepackage{caption}
\usepackage{float}
\usepackage{indentfirst}
\usepackage{mathtools}
\usepackage{algorithm, algorithmic}

\title{CLPSTNet: A Progressive Multi-Scale Convolutional Steganography Model Integrating Curriculum Learning
}
\author {
	FengChun Liu$^{1}$, Tong Zhang$^{2}$, Chunying Zhang$^{3}$\\
	$^{1}$Qianan College, North China University of Science and Technology, Tangshan, Hebei 063210, China\\
	$^{2}$School of Cyberspace Security, Beijing University of Posts and Telecommunications, Beijing 100876, China\\
	$^{3}$College of Science, North China University of Science and Technology, Tangshan, Hebei 063210, China\\
	\texttt{lnobliu@ncst.edu.cn, 2024010368@bupt.cn, hblg\_zcy@126.com}
}

\begin{document}
\maketitle

\begin{abstract}
In recent years, a large number of works have introduced Convolutional Neural Networks (CNNs) into image steganography, which transform traditional steganography methods such as hand-crafted features and prior knowledge design into steganography methods that neural networks autonomically learn information embedding. However, due to the inherent complexity of digital images, issues of invisibility and security persist when using CNN models for information embedding. In this paper, we propose Curriculum Learning Progressive Steganophy Network (CLPSTNet). The network consists of multiple progressive multi-scale convolutional modules that integrate Inception structures and dilated convolutions. The module contains multiple branching pathways, starting from a smaller convolutional kernel and dilatation rate, extracting the basic, local feature information from the feature map, and gradually expanding to the convolution with a larger convolutional kernel and dilatation rate for perceiving the feature information of a larger receptive field, so as to realize the multi-scale feature extraction from shallow to deep, and from fine to coarse, allowing the shallow secret information features to be refined in different fusion stages. The experimental results show that the proposed CLPSTNet not only has high PSNR , SSIM metrics and decoding accuracy on three large public datasets, ALASKA2, VOC2012 and ImageNet, but also the steganographic images generated by CLPSTNet have low steganalysis scores.You can find our code at \href{https://github.com/chaos-boops/CLPSTNet}{https://github.com/chaos-boops/CLPSTNet}.

\end{abstract}



\section{Introduction}\label{sec:intro}
With the digital development of Internet technology and information, people's lives can be convenient at the same time, the leakage of personal privacy, confidential data subject to malicious attacks and tampering and other problems are increasing, and the corresponding information security protection technology is developing rapidly. Information security protection technology mainly includes two kinds of information encryption and information hiding. Information encryption is a technology that utilizes a specific algorithm to transform the original plaintext information into unrecognizable ciphertext information, and then restores the ciphertext to plaintext through a decryption algorithm. Steganography is a technology that embeds secret information into a carrier in a covert way so as to protect the safe transmission of secret information. Steganography mainly researches how to embed secret information into other information carriers efficiently and securely, and conceal the existence of the information in the transmission process so as to guarantee the security of secret information transmission. Steganography embeds the secret information into the carrier through specific encoding algorithms, and then the receiver of the information realizes the extraction of the secret information through specific decoding algorithms.

Image steganography is the use of specific algorithms to embed secret information into digital images, and then use specific algorithms to extract the secret information from the digital images, generally used for the transmission of secret information, digital copyright authentication and other scenarios. For image steganography, the performance of an image steganography algorithm can be measured by embedding capacity, invisibility and security. With the introduction of deep learning into the field of steganography, image steganography models based on various types of neural network models have emerged in large numbers, greatly expanding the steganographic capacity of image steganography algorithms, but such deep learning image steganography algorithms perform poorly in terms of invisibility and security.

As a result, researchers have focused on how to improve the invisibility and security of deep learning image steganography, such as the use of adversarial training to improve security \cite{yang2023acgis}, improve the model structure to improve the invisibility \cite{tan2021channel} and so on. Adversarial training in image steganography algorithms originates from generative adversarial networks, where steganography and steganalysis networks are used as opposites, and the two are trained against each other to improve the steganographic image's resistance to steganalysis. The earliest SteGAN based on encoding-decoding network was proposed by Hayes et al \cite{hayes2017generating} for image steganography, which defines the tripartite adversarial of Alice, Bob and Eve, representing the process of image steganography - information extraction - steganalysis respectively, to improve the security of steganography algorithm.Zhang et al \cite{zhang2019steganogan} proposed SteganoGAN which includes the three-partite adversarial of encoding-decoding-evaluating parties, which has become the mainstream deep learning-based deep learning adversarial training. and becomes the current mainstream deep learning-based adversarial image steganography modeling framework. Firstly, the embedded information is transformed into binary data with tensor size of and spliced with the image in depth, the encoding network encodes it into the natural image with size of, the information is reconstructed from it by the decoding network, and the evaluation network is used to evaluate the performance of the encoding network in order to generate a more realistic cryptographic image.

Research for image steganography model structure focuses on improving the network's ability to capture multi-scale features \cite{zhang2019invisible}, mitigating the gradient vanishing problem \cite{duan2020high}, and new ways of embedding information \cite{jing2021hinet}. For example, SteganoGAN \cite{zhang2019steganogan} and FC-DenseNet\cite{duan2020high}, which choose dense connectivity module to mitigate the gradient vanishing problem; using fusion dilation convolution and dense connectivity module to extract multi-scale and multi-expansion rate image features \cite{wang2019hidinggan}, adaptively embedding more information in the redundant image region with rich texture; introducing Inception module \cite{zhang2019invisible} for fusing different sensory domain size feature maps to improve the network's ability to capture features at different scales. However, this type of model has the problems of color distortion and poor security in the generated steganographic images when embedding large amount of information.

The earliest curriculum learning method that appeared was a data-level sampling strategy, and with the gradual application of curriculum learning in various fields, the existing curriculum learning methods can be categorized into data-based, task-based, and network model-based curriculum learning according to the application object. Model-based curriculum learning algorithms make the network model obtain better performance by regularly transforming the network model during the training process. For example, gradually increasing the number of network layers \cite{karras2017progressive}, discarding neuron probabilities \cite{morerio2017curriculum}, etc. Inspired by curriculum learning algorithms, we propose an implicitly written network structure CLPSTNet (Curriculum Learning Progressive Steganography Network) that incorporates the idea of curriculum learning step-by-step, and design a progressive multi-scale convolution that incorporates the Inception structure and the dilated convolutional module. The shallow network starts from a convolution with a smaller convolution kernel and dilation rate, and gradually expands to a convolution with a larger convolution kernel and larger dilation rate, realizing feature extraction from shallow to deep and from fine to coarse. In summary, the main contributions of this paper include the following three parts:
\begin{itemize}
\item[1] A Progressive Multi-scale Convolution Block (PMCB), a multi-pathway convolution structure that incorporates Inception structure and dilation convolution, is designed to enhance the ability of the network to capture multi-scale features.
\item[2] A progressive multi-scale image steganography framework containing densely connection module and PMCB module is designed, which enables shallow information features to be refined at different fusion stages.
\item[3] Experiments on three large public datasets, ALASKA2, VOC2012, and ImageNet, show that the proposed steganography scheme has high steganographic quality metrics such as SSIM and PSNR, while the generated steganographic images have low steganalysis scores.
\end{itemize}

\section{Related Work}
In this section, we first present work on deep learning image steganography related to the study of model structure. In addition, we address curriculum learning algorithm concepts and related work.
\subsection{Deep learning steganography}

In 2014, GoodFellow et al \cite{goodfellow2020generative} proposed Generative Adversarial Networks (GAN), in which the generative and discriminative networks play with each other to produce a fairly good output. Since then, researchers have used Generative Adversarial Networks in the field of image steganography, and a large number of steganographic models based on various types of deep learning have emerged. The research on deep learning image steganography in terms of model structure improvement is mainly based on the introduction of multi-scale feature processing, the introduction of the attention mechanism, the introduction of the simulation of attack structure to improve the robustness and so on.

Research related to multi-scale feature processing mainly focuses on increasing the width of the network or introducing novel convolutions.Li et al \cite{wang2019hidinggan} proposed HCISNet, which fuses dilation convolution and dense connectivity modules to achieve the fusion of multi-scale and multi-dilatancy image features for better access to structurally and visually redundant regions of the image to enhance the steganographic capacity.Zhang et al \cite{zhang2019invisible} and Wang et al \cite{wang2019hidinggan} proposed to utilize the Inception module for fusing feature maps with different perceptual domain sizes to enhance the network's ability to capture features at different scales. Also by expanding the network width, the design of double convolution, including 1×1 convolution and 3×3 convolution, is added to the U-Net model in the study of Zeng et al \cite{zeng2023advanced}, which are all designed to enhance the network's adaptability to features of different scales by enhancing the network's width.Zhang et al \cite{zhang2023joint} propose cross-process comparative refinement of the CFCR and cross-process multiscale CPMS of the JAIS-Net for jointly adapting and refining images in the information hiding and recovery phases, where the differences between the current pairs of steganographic images at different scales are used to guide the next phase of the adapting and refining embedding process during the training process.

In order to enhance the invisibility of steganographic images, researchers have tried to introduce various types of attention mechanisms into the model structure improvement.Peng \cite{peng2024image}et al. proposed multi-scale channel attention to generate channel attention for feature maps at different scales, and orthogonal fusion was used to integrate the channel information from different sensory fields to improve the decoding accuracy. In Yao et al.'s \cite{yao2024high} study, multi-scale attention is generated by widening the network width, and branches with different convolutional kernel sizes are added to the base channel attention to obtain feature maps at different scales, which makes the model pay more attention to the features that are useful for improving the performance of information embedding. Swin Transformer has received much attention in the field of computer vision due to its excellent global modeling ability of multi-scale features from its self-attention module, and has likewise been applied in the field of image steganography.Kashif et al \cite{kashif2023enhanced} proposed enhanced pixel privacy utilizing Swin transformer with multiple heads of attention in combination with CycleGAN ( EPPGAN); Li et al \cite{li2023adversarial} proposed to add Shuffle Linear layer to the base of Swin transformer module to enhance the inductive bias ability of self-attention module, and to enhance the extraction of local features by shuffling the channels of the feature map in order to ensure that the feature information flows across the channels in the subsequent convolutional layers. Ke et al \cite{ke2024stegformer} proposed that the original self-attention module only considers spatial information and ignores channel information, and proposed Channel Adaptive Transformer Block (CATB) to utilize the global information of each channel to adjust the difference between channels.

The earliest reversible neural network for image steganography task was proposed by Jing et al \cite{jing2021hinet}, which explicitly modeled image recovery as an inverse process of image hiding, and only needed to train the network once to obtain all the parameters of the hiding and recovery networks, which achieved state-of-the-art performances in image recovery and hiding invisibility. Subsequently, Xu et al \cite{xu2022robust} proposed flow-based reversible neural network for steganography task, which is easier to compute and adds structures such as content-aware noise projection to improve the robustness of steganographic images. In addition, researchers have improved the invisibility of steganography by combining reversible neural networks with Swin transfomer \cite{feng2022image}, combined spatial channel attention with reversible neural networks to guide the embedding of secret information into more secure image regions \cite{li2023iscmis}, and utilized reversible neural networks to guide multi-image steganography \cite{guan2022deepmih}.

\subsection{Curriculum Learning}

The basic idea of curriculum learning originates from curriculum education in human behavior, humans need to undergo a long period of training from birth to adulthood, and this training is highly organized, with different concepts introduced at different stages, corresponding to a gradual increase in the difficulty, which leads to a gradual mastery of the knowledge learned. The concept of curriculum learning was initially proposed by Bengio et al \cite{bengio2009curriculum}, where an easier subset of data is used for training in the early stage of model training, and the difficulty of the subset of data is gradually increased until the entire data set is utilized for training, claiming that this makes it easy for the model to find better local optima, while speeding up the training speed. With the gradual application of curriculum learning in various fields, many research results have emerged. Existing curriculum learning methods can be categorized into data-based, task-based, and network model-based curriculum learning based on the object they are applied to \cite{liu2023review}.

Data-based curriculum learning advocates training models starting with simple samples and progressing gradually to complex samples. For example, in medical image analysis tasks, e.g., starting training from heavy disease images (the more severe the image lesion i.e., the simpler it is) \cite{tang2018attention} and progressively transitioning to moderate and mild, or starting training from images with nodules \cite{jesson2017cased} and unlabeled images containing high information content \cite{liu2022acpl} are used to balance out the problem of training bias in the medical image task because samples containing high information content have a higher probability of belonging to a minority class ( rare cases). Task-based curriculum learning approaches tasks incrementally by focusing on the connections between tasks, where each subtask is a simplified version of the next, and each task uses previously learned knowledge of the task. For example, starting with simpler tasks \cite{zhang2019curriculum}, more relevant \cite{pentina2015curriculum} task sets, etc., and gradually expanding to more difficult and less relevant tasks.

Model-based curriculum learning allows network models to achieve superior performance by regularly modifying them during the training process. For example, gradually increasing the number of network layers \cite{karras2017progressive}, controlling filters \cite{sinha2020curriculum}, discarding neuron probabilities \cite{morerio2017curriculum}, and increasing the capacity and strength of the discriminator \cite{kurmi2019curriculum,sharma2018improved}. In the research for generative adversarial network models, Karras et al\cite{karras2017progressive} used to start with a low resolution image, so that the model captures the contour information of the data from it, and gradually add new network layers dealing with higher resolution details during the subsequent training process, which are used to increase the detail information of the image; Sharma et al \cite{szegedy2015going} proposed that by continuously enhancing the discriminator's discriminatory ability is used to find the generator the problem that the generator needs to progress under increasingly difficult curriculum tasks in order to deceive the discriminator and achieve high quality images.

\section{CLPSTNet}
In this section, we present the overall architecture of CLPSTNet, describing in detail the design of the progressive multiscale convolutional module, followed by the structure of the encoding and decoding networks, and the cryptographic analysis network. Finally, the definition of the loss function is given.
\subsection{Overall framework}

\begin{figure}[h]
	\centering
	\centering
	\includegraphics[width=1\linewidth]{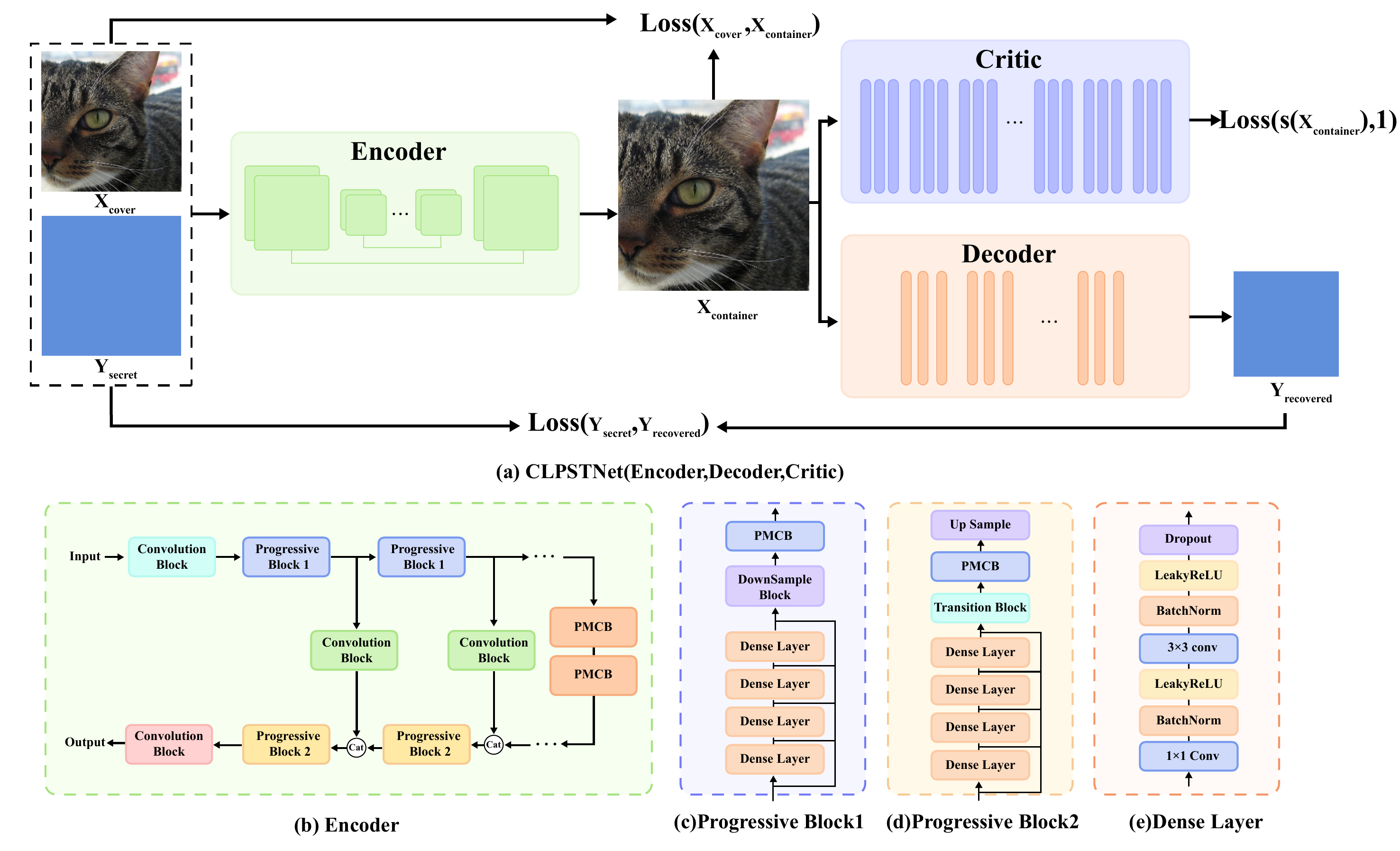}
	\caption{Framework of CLPSTNet}
	\label{fig1}
\end{figure}

As shown in Figure \ref{fig1}, the overall framework of CLPSTNet includes encoding network, decoding network and steganalysis network.
The encoding network accomplishes information hiding, the decoding network accomplishes information recovery, and the steganography analysis network is used to evaluate the performance of the encoding network. Encoding network Encoder takes original image and secret information as input. 

Assuming that the inputs are the original image $X_{cover} \in R^{3 \times H \times W}$ and the secret information $Y_{secret} \in R^{D \times H \times W}$, the original image and the secret information are spliced on the channel as inputs to the encoding network $I \in R^{(3+D) \times H \times W}$:

\begin{equation}
I = Cat(X_{cover}, Y_{secret})
\end{equation}

After processing by the Encoder, the steganographic image $X_{container} \in R^{3 \times H \times W}$ is generated, and the process can be described as:
\begin{equation}
Encoder : (X_{cover}, Y_{secret}) \rightarrow X_{container} 
\end{equation}

The decoding network Decoder accepts the steganographic image as input and generates the recovered secret message $Y_{recovered} \in R^{D \times W \times H}$ from the steganographic image, the process can be described as:
\begin{equation}
Decoder : X_{container} \rightarrow Y_{recovered}
\end{equation}
The main goal of the coding network is to create a steganographic image that keeps the visual appearance of the original image intact while preserving the secret information in the steganographic image from being discovered. The steganalysis network Critic takes the steganographic image and the original image as inputs and outputs a steganalysis score $s \in [0,1]$. When the score is closer to 1, it indicates that the image is more likely to contain secret information:

\begin{equation}
Critic : (X_{cover}, X_{container}) \rightarrow s 
\end{equation}

\subsection{PMCB Module}

\begin{figure}[h]
	\centering
	\centering
	\includegraphics[width=0.7\linewidth]{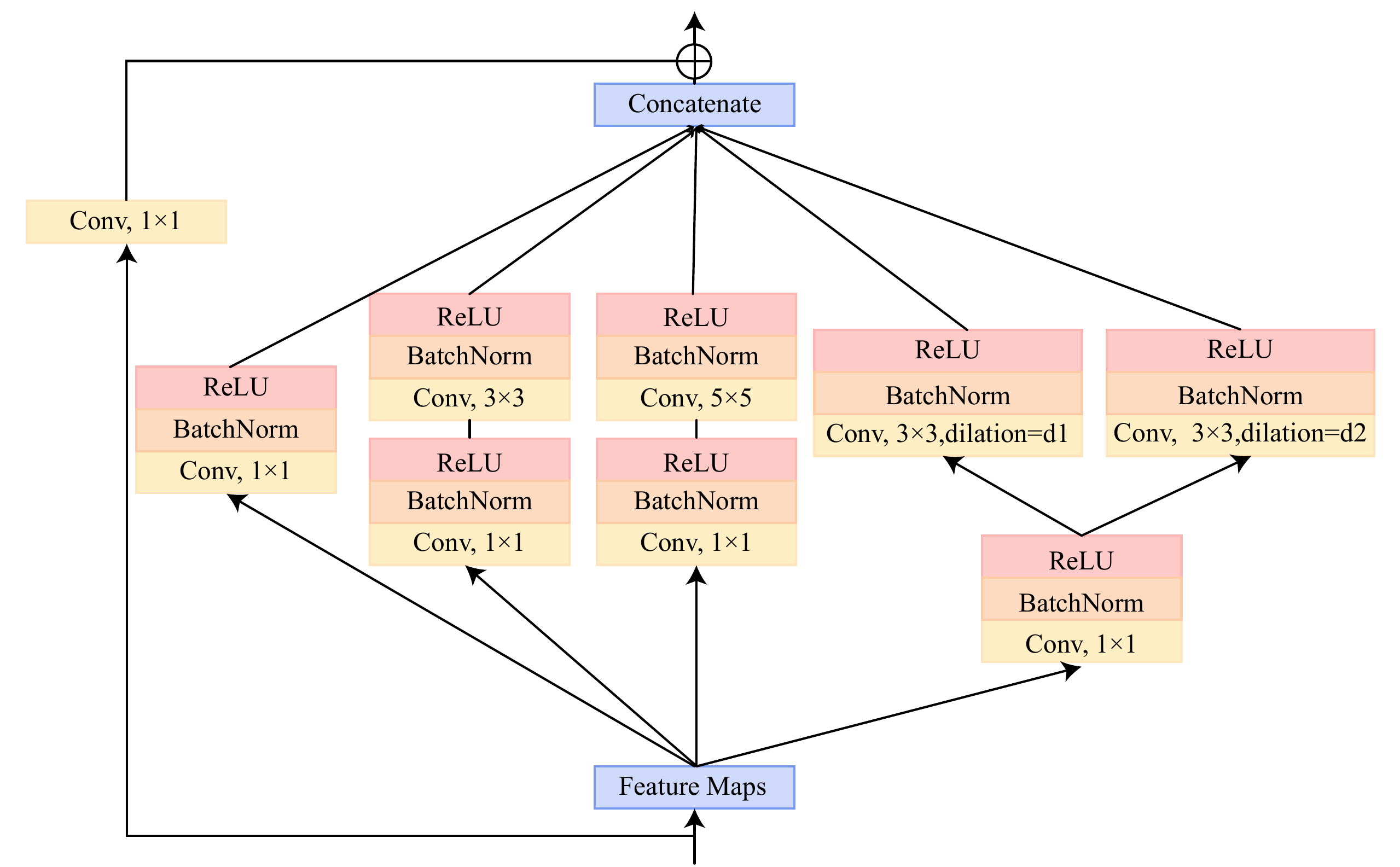}
	\caption{PMCB Module}
	\label{fig2}
\end{figure}

In the process of realizing the embedding and reconstruction of multi-channel secret information, the quality of the steganographic image and the reconstruction decoding rate decrease with the expansion of the embedding capacity, which is related to the inability of the encoding and decoding network to deal with feature information of different scales. Inspired by curriculum learning, it mimics the process of human learning, starting from smaller and simpler knowledge and gradually expanding to macroscopic and more complex knowledge. As shown in Figure \ref{fig2}.

Fusing the Inception structure and dilation convolution to construct a multi-pathway convolution structure, the Progressive Multi-scale Convolution Block (PMCB) is constructed, starting from a convolution containing a smaller convolution kernel and a dilation rate, to extract the most basic and localized feature information from the feature map. As the network deepens, it gradually expands to a convolution containing a larger convolution kernel and dilatation rate for sensing feature information of a larger range and larger sensory field, and extracts features from shallow to deep and from fine to coarse, with each layer observing different information features for each different convolution pathway to enhance the model's ability of extracting multi-scale information features.

A single PMCB module contains five branching pathways, including 1×1conv, 3×3conv, 5×5conv, 3×3 dilated convolution with dilation rate D1, and 3×3 dilated convolution with dilation rate D2. The 3×3 convolution kernel and 5×5 convolution kernel, and the two dilation convolution branching paths are preceded by 1×1conv added for dimensionality reduction to reduce the computational and parametric quantities due to the addition of large convolutions, and to increase the model's nonlinear expressive capability. The feature maps of the previous layer are input to the PMCB module, and are processed in parallel by the five branching paths, thus obtaining the multiscale coded information, and finally the results of the five branching paths are channel spliced in order to fuse the multiscale information. Compared with the original Inception module, the PMCB module adds the branching path of dilation convolution, which introduces dilation convolution processing to increase the receptive field and obtain denser feature information, which is especially important for image steganography to better obtain the structural and visual redundancy regions of the image to enhance the steganographic capacity.

Progressive Multiscale Convolution Module PMCB The progressive growth of scale in the coding network is reflected in the size of the convolution kernel, dilation rate. First, the shallow layer of the network starts after the PMCB1 module containing 1×1conv, 3×3conv, 5×5conv, dilation rate (Dilation=3) dilation convolution, and dilation rate (Dilation=6) dilation convolution, from which the network captures local details of a smaller scale and a smaller perceptual range, and the shallow network focuses more on the edge and detail features. It then expands to the PMCB2 module containing 1×1conv, 3×3conv, 5×5conv, dilation rate (Dilation=6) dilation convolution, and dilation rate (Dilation=12), which gradually captures feature information with larger sensory fields and more scales, and finally contains 1×1conv, 3×3conv, 5×5conv, dilation rate ( Dilation=12) dilation convolution, and PMCB3 module with dilation rate (Dilation=18) to perceive feature information with larger sensory field and more scales, so that the secret information features and image information can be fused and refined at different fusion stages through the processing of PMCB modules with different scales, in order to enhance the feature processing capability of the network for the fusion and separation of secret information and images.

\subsection{Encoder}

As shown in Figure \ref{fig1}(b), the complete Encoder model mainly contains an initial convolution block, a downsampling stage, an upsampling stage and an output convolution block. First, the input of Encoder is processed by the initial convolution module, which contains four residual convolution modules, and the resolution of the output feature map is not changed. Each residual convolution module contains 3×3 conv, BatchNorm and LeakyReLU and a residual short-circuit join. Next, the feature maps here are processed by three Progressive Blocks1, each of which contains four layers of densely connected layers, downsampling layers, and PMCB modules, as shown in Figure \ref{fig1}(c).

The structure of each densely connected layer is shown in Figure \ref{fig1}(e), containing 1 × 1 conv, BatchNorm, LeakyReLU and Dropout, and the output of each layer is added to the subsequent network layer by channel splicing. The inputs of each layer consider the feature mapping of all previous layers. The dense connectivity module is used to alleviate the problem of gradient vanishing due to deepening of the network and to enhance feature propagation to realize the reuse of low-level feature information about edges, shapes, etc. in the shallow layer, so that the shallow secret information features and image features flow in different layers of the network. The resolution of the feature map output inside the dense connectivity module is kept constant.

The downsampling layer contains a 3×3 conv, BN, activation function LeakyReLU with a step size of 2. The resolution of the feature map output from this layer is halved. The internal construction in each PMCB module during the encoding process is not always the same, where the two branching paths of the dilation convolution use progressive growth, and the dilation rate parameter for this part uses [3,6,12,18,3,6,12,18], which is followed to ensure that the region of use under the sensory field is continuous. The output of each Progressive Block1 is fed individually to the Jump Connection Module for processing and is used to recover feature map resolution and supplemental details.

Subsequently, two PMCB modules are used as bottleneck layers for capturing multi-scale feature information at low resolution. Next, we use a U-Net structure similar to downsampling for feature reconstruction, and the decoding stage also contains three Progressive Blocks2, as shown in Figure \ref{fig1}(d). Each Progressive Block2 contains four layers of dense connectivity layer, PMCB module, Transition Block and upsampling layer. Each Progressive Block2 in the decoding stage accepts the processed feature maps from each Progressive Block1 in the encoding stage, and performs feature fusion by channel splicing after changing the number of channels via 1×1 conv, which is used to recover the feature map resolution and complementary details. The feature maps are then processed by the Dense Linking Module without changing the resolution of the feature maps. Then it is processed by the Transition Block module for adjusting the number of channels and reducing the computational effort, which contains BN, LeakyReLU, and 1×1conv. Subsequently it is processed by the PMCB module with different dilation rate settings, and the input is fed into the upsampling module, which contains an inverse convolutional module, BN, and LeakyReLU for recovering the feature map resolution. Finally after two residual convolution layers are used for output.

\subsection{Decoder}
Decoder accepts the output of Encoder as input and recovers the secret information from the steganographic image.Decoder consists of initial convolutional block, PMCB module and output convolutional block, the specific network structure and the shape of output tensor are shown in Table \ref{tab1}.

\begin{table}
	\centering
	\caption{Decoder model structure}
	\renewcommand{\arraystretch}{1.3}
	\label{tab1}  
	\begin{tabular}{ccc}
		\hline
		Group/Layer & Progress & Output size \\
		\hline
		Input & / & $128\times128\times3$ \\
		Residual Block & Conv+BN+ReLU & $128\times128\times32$ \\
		PMCB & Dilation=(3,6) & $128\times128\times192$ \\
		PMCB & Dilation=(6,12) & $128\times128\times288$ \\
		PMCB & Dilation=(12,18) & $128\times128\times512$ \\
		Conv & / & $128\times128\times D$ \\
		\hline
	\end{tabular}
\end{table}

\subsection{Critic}

Critic accepts the output of Encoder and the original image together as input, and outputs a score representing the probability that the image contains secret information, the closer the score is to 1, i.e., the higher the probability that the image contains secret information, which means that the image is more likely to be detected by the steganalysis network.Critic chooses the XuNet-based steganalysis auxiliary network proposed by Zhang et al \cite{zhang2019invisible}. Critic uses XuNet based steganalysis assisted network proposed by Zhang et al \cite{zhang2019invisible}. The steganalysis network is used to evaluate the steganalysis resistance of the samples during the training process, and the encoder network-decoder network is formed with adversarial training to improve the steganalysis resistance of the encrypted image, the specific network structure and the shape of the output tensor of each layer are shown in Table \ref{tab2}.

\begin{table}
	\centering
	\caption{Critic model structure}
	\renewcommand{\arraystretch}{1.3}
	\label{tab2}  
	\begin{tabular}{ccc}
		\hline
		Group/Layer & Progress & Output size \\
		\hline
		Input & / & $128\times128\times3$ \\
		ConvBlock1 & Conv+BN+ReLU+AvgPool & $64\times64\times8$ \\
		ConvBlock1 & Conv+BN+Tanh+AvgPool & $32\times32\times16$ \\
		ConvBlock2 & Conv+BN+ReLU+AvgPool & $16\times16\times32$ \\
		ConvBlock2 & Conv+BN+ReLU+AvgPool & $8\times8\times64$ \\
		ConvBlock3 & Conv+BN+ReLU & $4\times4\times128$ \\
		SPPBlock & / & $3840\times1$ \\
		FC & / & $128\times1$ \\
		FC & / & $2\times1$ \\
		\hline
	\end{tabular}
\end{table}

\subsection{Loss function}

In order to improve the performance of the steganographic model, the objective function of the CLPSTNet model can be categorized into embedding loss, recovery loss and steganalysis loss, which can be defined as follows:
\begin{equation}
L_{total} = L_{encode} + a \times L_{decode} + b \times L_{stehsis} 
\end{equation}
where $L_{total}$ is the overall loss function of the network, $L_{encode}$ is the embedding loss for information hiding, $L_{decode}$ is the recovery loss for information reconstruction, and $L_{stehsis}$ is the loss for steganalysis. $\alpha, \beta$ are the parameters used to balance the three components of the loss.

Embedding Loss: In order to ensure the visual consistency between the original image and the steganographic image, we consider several metrics for evaluating image similarity, including Mean Square Error (MSE), Structure Similarity Index Measure (SSIM), and Multi-Scale Structure Similarity Index ( Multi-Scale Structure Similarity Index Measure, MSSSIM)\cite{34}
\begin{equation}
L_{encode} = \lambda_1 \times MSE(x, y) + \lambda_2 \times (1-SSIM(x, y)) + \lambda_3 \times (1-MSSSIM(x, y)) 
\end{equation}
$\lambda_1\lambda_2\lambda_3$ are the parameters used to balance the three components of the embedding loss. MSE is chosen to measure the pixel-level difference between the original image and the steganographic image pair, and SSIM and MSSSIM are chosen to ensure the visual consistency of the image pairs on the human visual system.SSIM takes into account the sensitivity property of the human visual system to the structural information in terms of brightness, contrast and structural information of the images.SSIM takes the value of [0,1], and when the value of SSIM is closer to 1 The value of SSIM is [0,1], when SSIM value is closer to 1, it indicates that the two images are more similar in terms of brightness, contrast and structural information:
\begin{equation}
SSIM(x, y) = \frac{(2\mu_x\mu_y) + C_1}{{\mu_x}^2 + {\mu_y}^2 + C_1}^\alpha \cdot \frac{2\sigma_x\sigma_y + C_2}{{\sigma_x}^2 + {\sigma_y}^2 + C_ 2}^\beta \cdot \frac{\sigma_{xy} + C_3}{\sigma_x\sigma_y + C_3}^\gamma 
\end{equation}
where $\mu_x, \mu_y$ are the mean values of the original and loaded images respectively, $\sigma_x, \sigma_y$ is the variance of the original and loaded images respectively, $\sigma_{xy}$ is the covariance of the original and loaded images, $C_1, C_2, C_3$ is used to avoid parameters with denominators close to zero, $\alpha > 0, \beta > 0, \gamma > 0$ which are parameters used to adjust the importance of the three components.

MSSSIM is a variant of SSIM, a multiscale based SSIM metric that iteratively downsamples the image using a low-pass filter, with the original image having a scale of 1 and the highest scale of M. MSSSIM is obtained by calculating on different scales:
\begin{equation}
MSSSIM(x,y) = [l(x,y)]^{\alpha M} \times \prod_{j=1}^{M} [c_j(x,y)]^{\beta j} \times [s_j(x,y)]^{\gamma j} 
\end{equation}
Recovery loss: Binary cross entropy (BCE) is chosen for evaluating the accuracy of information recovery:
\begin{equation}
L_{decode} = BCE(Y_{secret}, Y_{recovered}) 
\end{equation}

\section{Experiments}

\subsection{Experimental Platform and Datasets}

The experiments in this paper were all conducted on a Linux operating system using the PyTorch 1.10.1 deep learning framework, with the system's GPU being the NVIDIA GeForce RTX 3090. 
The datasets used were the ALASKA2, Pascal VOC2012, and ImageNet, which are three publicly available large-scale datasets. The ALASKA2 is a public dataset from the ALASKA2 Image Steganalysis competition on the Kaggle platform. Within the ALASKA2 dataset, 10,000 original images from the "Cover" category were selected for the training set, 3,000 for the validation set, and 7,000 for the test set. ImageNet is a large-scale public computer vision dataset; 25,000 images were extracted from it, with 20,000 used for the training set and the remainder for testing. VOC2012 is a dataset used for object detection and semantic segmentation; 13,000 images were selected from it to form the training set, with the remaining 5,000 used for the test and validation sets. Due to computational power limitations, all original images from the datasets were processed through a Matlab program to be resized to 128×128 pixels.

\subsection{Parameters}
The experiments in this paper utilize the Adaptive Moment Estimation (Adam) algorithm provided by the PyTorch platform to optimize the encoding and decoding networks. The initial learning rate is set to 0.001, with momentum parameters (betas) configured as (0.9, 0.999). The Stochastic Gradient Descent (SGD) algorithm is selected to optimize the steganalysis adversarial network, with the initial learning rate (lr) set to 0.0001/3, and weight decay (weight decay) set to 1e-8, updating the steganalysis network parameters every 5 batches. The number of samples selected for each training session (Batch size) is 8, and the maximum number of iterations for the model (max-iter) is 120. 
The loss function comprises encoding loss, decoding loss, and steganalysis loss. The encoding loss incorporates the Structural Similarity Index (SSIM), Multi-Scale Structural Similarity (MS-SSIM), and Mean Square Error (MSE) as evaluation metrics, with corresponding proportional coefficients of 0.5:0.5:0.3. Both the decoding loss and the steganalysis loss employ binary cross-entropy. The ratio of the encoding loss to decoding loss to steganalysis  loss in the loss function is 1:1:0.1. In the experiments, the steganographic capacity is D=1-6 bpp (that is, the hidden tensor in a 128×128 image is 128×128×D in size).

\subsection{Experimental results}
1) Test results of the model on multiple datasets. The CLPSTNet model is selected to be tested on the 1-6 bpp ALASKA2, VOC2012, and ImageNet datasets, and the experimental results are shown in Table 3. As can be seen from Table \ref{tab3}, the CLPSTNet model performs well in all indicators of 1-6 bpp steganographic capacity on the two datasets, with the PSNR exceeding 40 under the 1-3 bpp indicator, and the SSIM indicator is close to 1, which indicates that the steganographic image generated by the CLPSTNet model is very similar to the original image.

\begin{table}
	\centering
	\caption{Performance of the CLPSTNet model in steganography and decoding accuracy}
	\label{tab3}
	\renewcommand{\arraystretch}{1.3}
	\begin{tabular}{ccccccc}
		\hline
		Dataset & D & SSIM & MSSSIM & PSNR & RMSE & Accuracy \\
		\hline
		\multirow{6}{*}{ALASKA2} 
		& 1 & 0.99932 & 0.99997 & 50.983 & 0.0028 & 0.98 \\
		& 2 & 0.99889 & 0.99996 & 44.819 & 0.0059 & 0.87 \\
		& 3 & 0.99518 & 0.99970 & 42.098 & 0.0092 & 0.60 \\
		& 4 & 0.98896 & 0.99920 & 39.190 & 0.0116 & 0.66 \\
		& 5 & 0.99184 & 0.99917 & 38.959 & 0.0319 & 0.67 \\
		& 6 & 0.98991 & 0.99919 & 37.419 & 0.0162 & 0.64 \\
		\hline
		\multirow{6}{*}{VOC2012} 
		& 1 & 0.99930 & 0.99994 & 50.215 & 0.0032 & 0.94 \\
		& 2 & 0.99840 & 0.99993 & 46.085 & 0.0050 & 0.96 \\
		& 3 & 0.99550 & 0.99946 & 38.554 & 0.0132 & 0.68 \\
		& 4 & 0.99049 & 0.99943 & 39.399 & 0.0109 & 0.62 \\
		& 5 & 0.98053 & 0.99730 & 34.065 & 0.0249 & 0.72 \\
		& 6 & 0.96507 & 0.99732 & 33.874 & 0.0312 & 0.75 \\
		\hline
		\multirow{6}{*}{ImageNet} 
		& 1 & 0.99933 & 0.99997 & 49.537 & 0.0034 & 0.95 \\
		& 2 & 0.99862 & 0.99991 & 44.573 & 0.0062 & 0.83 \\
		& 3 & 0.99819 & 0.99989 & 43.884 & 0.0067 & 0.83 \\
		& 4 & 0.99486 & 0.99971 & 37.994 & 0.0147 & 0.65 \\
		& 5 & 0.98880 & 0.99925 & 37.000 & 0.0207 & 0.67 \\
		& 6 & 0.98810 & 0.99887 & 36.669 & 0.0178 & 0.66 \\
		\hline
	\end{tabular}
\end{table}

2) PMCB module validation for coding and decoding effectiveness. In order to explore the effectiveness of PMCB module in encoding and decoding networks, the Conv model, which contains only the base convolutional module, is chosen as the baseline model, and the ProgressiveNet model, which contains only the PMCB module in the decoding network and only the base convolutional module in the encoding network, is chosen as a comparative model, to be compared with the CLPSTNet model, which contains the PMCB module in both the encoding and decoding networks. CLPSTNet model for comparison. The 1-6 bpp steganographic capacity on the ALASKA2 dataset was chosen for the experiments, and the experimental results are shown in Table \ref{tab4}.

\begin{table}[htbp]
	\centering
	\caption{Impact of the PMCB Module on the Performance of the Encoding and Decoding}
	\label{tab4}
	\renewcommand{\arraystretch}{1.3}
	\begin{tabular}{cccccc}
		\hline
		D & Model & SSIM & MSSSIM & PSNR & Accuracy \\
		\hline
		\multirow{3}{*}{1} 
		& Conv & 0.99027 & 0.99806 & 34.260 & 0.74 \\
		& ProgressiveNet & 0.98063 & 0.99771 & 33.788 & \textbf{0.99} \\
		& CLPSTNet & \textbf{0.99932} & \textbf{0.99997} & \textbf{50.983} & 0.98 \\
		\hline
		\multirow{3}{*}{2} 
		& Conv & 0.98694 & 0.99651 & 35.263 & 0.61 \\
		& ProgressiveNet & 0.98934 & 0.99847 & 34.324 & \textbf{0.99} \\
		& CLPSTNet & \textbf{0.99889} & \textbf{0.99996} & \textbf{44.819} & 0.87 \\
		\hline
		\multirow{3}{*}{3} 
		& Conv & 0.98632 & 0.99714 & 34.183 & 0.57 \\
		& ProgressiveNet & 0.99489 & 0.99877 & 35.714 & \textbf{0.87} \\
		& CLPSTNet & \textbf{0.99518} & \textbf{0.99970} & \textbf{42.098} & 0.60 \\
		\hline
		\multirow{3}{*}{4} 
		& Conv & 0.98452 & 0.99703 & 33.768 & 0.52 \\
		& ProgressiveNet & \textbf{0.99267} & 0.99715 & 33.334 & \textbf{0.70} \\
		& CLPSTNet & 0.98996 & \textbf{0.99920} & \textbf{39.190} & 0.66 \\
		\hline
		\multirow{3}{*}{5} 
		& Conv & 0.97835 & 0.99329 & 31.010 & 0.52 \\
		& ProgressiveNet & 0.99136 & 0.99760 & 34.427 & 0.67 \\
		& CLPSTNet & \textbf{0.99184 }& \textbf{0.99917} & \textbf{38.959} & 0.67 \\
		\hline
		\multirow{3}{*}{6} 
		& Conv & 0.97987 & 0.99405 & 32.366 & 0.50 \\
		& ProgressiveNet & 0.98845 & 0.99733 & 34.384 & 0.64 \\
		& CLPSTNet & \textbf{0.98991} & \textbf{0.99919} & \textbf{37.419} & 0.64 \\
		\hline
	\end{tabular}
\end{table}
As can be seen from Table \ref{tab4}, the CLPSTNet model outperforms the baseline model in 1-6 bpp steganographic capacity, and outperforms the baseline model in several metrics of steganographic image quality, such as SSIM, PSNR, and MSSSIM, while the decoding accuracy is the same or slightly lower than that of the baseline model. On the other hand, ProgressiveNet with PMCB selected for decoding network and basic convolutional structure selected for encoding network has lower performance than CLPSTNet in several metrics of steganographic image quality, while the decoding accuracy is slightly higher than that of CLPSTNet and the baseline model. It shows that the PMCB module is very effective in improving the decoding accuracy, while the CLPSTNet model, which adopts the PMCB module for both the encoding and decoding networks, is higher than PregressiveNet and the baseline model in several metrics of the steganographic quality assessment, indicating that the PMCB module is also effective in improving the quality of steganographic images.

3) PMCB module and Inception module comparison experiment. In order to further explore the effects of the basic Inception module and the PMCB module improved by the Inception module on the model performance, the steganographic network containing only the Inception module, the steganographic network containing the Inception module and the dense connectivity module, and the CLPSTNet containing the PMCB module and the dense connectivity module were selected for comparison. The 1-3 bpp steganographic capacity on the ALASKA2 dataset was chosen for the experiments, and the experimental results are shown in Table \ref{tab5}.

\begin{table}[htbp]
	\centering
	\caption{Comparative experimental results of PMCB module and Inception module}
	\label{tab5}
	\renewcommand{\arraystretch}{1.3}
	\begin{tabular}{ccccccccc}
		\hline
		D & Inception & PMCB & Dense Block & SSIM & MSSSIM & PSNR & RMSE & Accuracy \\
		\hline
		\multirow{3}{*}{1} 
		& $\sqrt{}$ & $\times$ & $\times$ & 0.98312 & 0.99625 & 28.286 & 0.0386 & 0.63 \\
		& $\sqrt{}$ & $\times$ & $\sqrt{}$ & 0.99850 & 0.99992 & 47.842 & 0.0040 & 0.79 \\
		& $\times$ & $\sqrt{}$ & $\sqrt{}$ & \textbf{0.99932} & \textbf{0.99997} & \textbf{50.983} & \textbf{0.0028} & \textbf{0.98} \\
		\hline
		\multirow{3}{*}{2} 
		& $\sqrt{}$ & $\times$ & $\times$ & 0.95990 & 0.99114 & 27.974 & 0.0400 & 0.64 \\
		& $\sqrt{}$ & $\times$ & $\sqrt{}$ & 0.99316 & 0.99962 & 40.577 & 0.0094 & 0.82 \\
		& $\times$ & $\sqrt{}$ & $\sqrt{}$ & \textbf{0.99889} & \textbf{0.99996} & \textbf{44.819} & \textbf{0.0059} & \textbf{0.87} \\
		\hline
		\multirow{3}{*}{3} 
		& $\sqrt{}$ & $\times$ & $\times$ & 0.92621 & 0.97698 & 23.670 & 0.0716 & 0.50 \\
		& $\sqrt{}$ & $\times$ & $\sqrt{}$ & 0.97997 & 0.99915 & 39.821 & 0.0110 & 0.50 \\
		& $\times$ & $\sqrt{}$ & $\sqrt{}$ & \textbf{0.99518} & \textbf{0.99970} & \textbf{42.098} & \textbf{0.0092} & \textbf{0.60} \\
		\hline
	\end{tabular}
\end{table}

As can be seen from Table \ref{tab5}, the PMCB module improved by the Inception module outperforms the model containing only Incepiton or the model containing the Incetpion module and the Dense Connection module in several metrics of implicit writing evaluation, such as SSIM, PSNR, and Decoding Accuracy, indicating that the PMCB module is more effective in improving the model performance compared to the Inception module.

4) PMCB module and dense connectivity module validation. In order to investigate the effect of the internal structure of the proposed CLPSTNet model on the model performance, the dense connectivity module and the PMCB module, which are included in the coding network of the CLPSTNet model, are selected for ablation experiments. The test results of 1-3 bpp hidden writing capacity on the dataset ALASKA2 are shown in \ref{tab6}.

\begin{table}
	\centering
	\caption{Influence of densely connected module and PMCB module on model performance}
	\label{tab6}
	\renewcommand{\arraystretch}{1.3}
	\begin{tabular}{ccccccc}
		\hline
		D & Dense Block & PMCB Block & SSIM & MSSSIM & PSNR & Accuracy \\
		\hline
		\multirow{4}{*}{1} 
		& $\times$ & $\times$ & 0.99027 & 0.99806 & 34.260 & 0.74 \\
		& $\sqrt{}$ & $\times$ & 0.99918 & \textbf{0.99997} & 46.299 & 0.95 \\
		& $\times$ & $\sqrt{}$ & 0.99916 & 0.99995 & 43.728 & 0.83 \\
		& $\sqrt{}$ & $\sqrt{}$ & \textbf{0.99932} & \textbf{0.99997} & \textbf{50.983} & \textbf{0.98} \\
		\hline
		\multirow{4}{*}{2} 
		& $\times$ & $\times$ & 0.98694 & 0.99651 & 35.263 & 0.61 \\
		& $\sqrt{}$ & $\times$ & 0.99839 & 0.99994 & \textbf{47.565} & 0.82 \\
		& $\times$ & $\sqrt{}$ & 0.99682 & 0.99972 & 39.228 & 0.81 \\
		& $\sqrt{}$ & $\sqrt{}$ & \textbf{0.99889} & \textbf{0.99996} & 44.819 & \textbf{0.87} \\
		\hline
		\multirow{4}{*}{3} 
		& $\times$ & $\times$ & 0.98632 & 0.99714 & 34.183 & 0.57 \\
		& $\sqrt{}$ & $\times$ & 0.99583 & 0.99979 & \textbf{43.569} & 0.69 \\
		& $\times$ & $\sqrt{}$ & \textbf{0.99784} & \textbf{0.99985} & 42.897 & \textbf{0.83} \\
		& $\sqrt{}$ & $\sqrt{}$ & 0.99518 & 0.99970 & 42.098 & 0.60 \\
		\hline
	\end{tabular}
\end{table}

As can be seen from Table \ref{tab6}, the CLPSTNet models containing only the dense connectivity module, only the PMCB module, and both the PMCB module and the dense connectivity module outperform the baseline model in terms of image steganography quality and decoding accuracy, indicating the effectiveness of the dense connectivity module and the PMCB module. At 1 bpp steganographic capacity, the CLPSTNet model outperforms the rest of the compared models in all cases, indicating that the proposed CLPSTNet model structure has excellent steganographic performance and decoding accuracy.

5) Comparison of CLPSTNet progressive modules.The expansion rate of several PMCB modules in the CLPSTNet network is set to progressive growth.In order to explore the effectiveness of this progressive idea, the schemes that contain only the expansion rate of 3, 6, 3, and 6 parameters are selected to be compared, and the test results are shown in Table \ref{tab7}. As can be seen from Table 7, the scheme with the expansion rate parameter of the PMCB module set to progressive growth performs the best in all the indicators, including SSIM, PSNR and decoding accuracy, which are all better than the PMCB module scheme with only a single expansion rate set, indicating the effectiveness of the progressive PMCB module in CLPSTNet.

\begin{table}
	\centering
	\caption{Influence of dilation rate parameters in PMCB module on model performance}
	\label{tab7}
	\renewcommand{\arraystretch}{1.3}
	\begin{tabular}{ccccccc}
		\hline
		Dilation & SSIM & MSSSIM & PSNR & RMSE & Accuracy \\
		\hline
		(3) & 0.99941 & 0.99996 & 48.855 & 0.0036 & 0.94 \\
		(6) & 0.99918 & 0.99996 & 48.991 & 0.0036 & 0.89 \\
		(3,6) & \textbf{0.99948} & \textbf{0.99998} & 50.511 & 0.0030 & 0.93 \\
		CLPSTNet & 0.99932 & 0.99997 & \textbf{50.983} & \textbf{0.0028} & \textbf{0.98} \\
		\hline
	\end{tabular}
\end{table}

6) Comparison with other models. In order to more fully demonstrate the excellent performance of CLPSTNet, different structures of image steganography models are selected for experimental comparative analysis, including the classical image steganography network structures ResNet \cite{he2016deep}, SteganoGAN \cite{zhang2019steganogan}, and HCISNet \cite{wang2019hidinggan}, as well as the multiscale network models DenseASPP \cite{yang2018denseaspp}, FC-DenseNet \cite{duan2020high}, the as well as network structures that include attention mechanisms such as SENet \cite{hu2018squeeze}, ECANe \cite{wang2020eca} and CBAM \cite{woo2018cbam}. In addition, ProgressiveNet models that only decode networks containing PMCB modules were added for comparison, and the test results of these models and the CLPSTNet model for 1bpp hidden write capacity on the dataset ALASKA2 are shown in Table \ref{tab8}.

\begin{table}
	\centering
	\caption{Comparison experiments between CLPSTNet and other models}
	\label{tab8}
	\renewcommand{\arraystretch}{1.3}
	\begin{tabular}{cccccc}
		\hline
		Model & SSIM & MSSSIM & PSNR & RMSE & Accuracy \\
		\hline
		Conv & 0.98351 & 0.99771 & 33.788 & 0.020 & \textbf{0.99} \\
		ResNet & 0.98993 & 0.99708 & 33.435 & 0.021 & 0.78 \\
		steganGAN & 0.98970 & 0.99907 & 37.669 & - & 0.95 \\
		DenseASPP & 0.99753 & 0.99933 & 36.008 & - & 0.95 \\
		SENet & 0.99432 & 0.99881 & 37.131 & 0.014 & 0.96 \\
		ECANet & 0.99438 & 0.99919 & 39.934 & 0.010 & 0.86 \\
		CBAM & 0.99470 & 0.99869 & 37.878 & 0.012 & 0.92 \\
		FC-DenseNet & 0.98179 & 0.99236 & 30.510 & - & 0.72 \\
		HCISNet & 0.96532 & 0.99822 & 41.221 & - &\textbf{0.99} \\
		ProgressiveNet & 0.98351 & 0.99771 & 33.788 & 0.020 &\textbf{0.99} \\
		CLPSTNet & \textbf{0.99932} & \textbf{0.99997} & \textbf{50.983} & \textbf{0.0028} & 0.98 \\
		\hline
	\end{tabular}
\end{table}

As can be seen from Table \ref{tab8}, the steganographic quality metrics MSSSIM and PSNR of CLPSTNet are higher than those of the other comparison models, which proves the excellent performance of the CLPSTNet model in image steganography quality. Under the same experimental conditions, the decoding accuracy of the ProgressiveNet model with only the decoding network containing the PMCB module has the best performance among the many comparison models, indicating the effectiveness of the PMCB module in improving the decoding accuracy.

\begin{figure}[!htb]
	\centering
	\centering
	\includegraphics[width=1\linewidth]{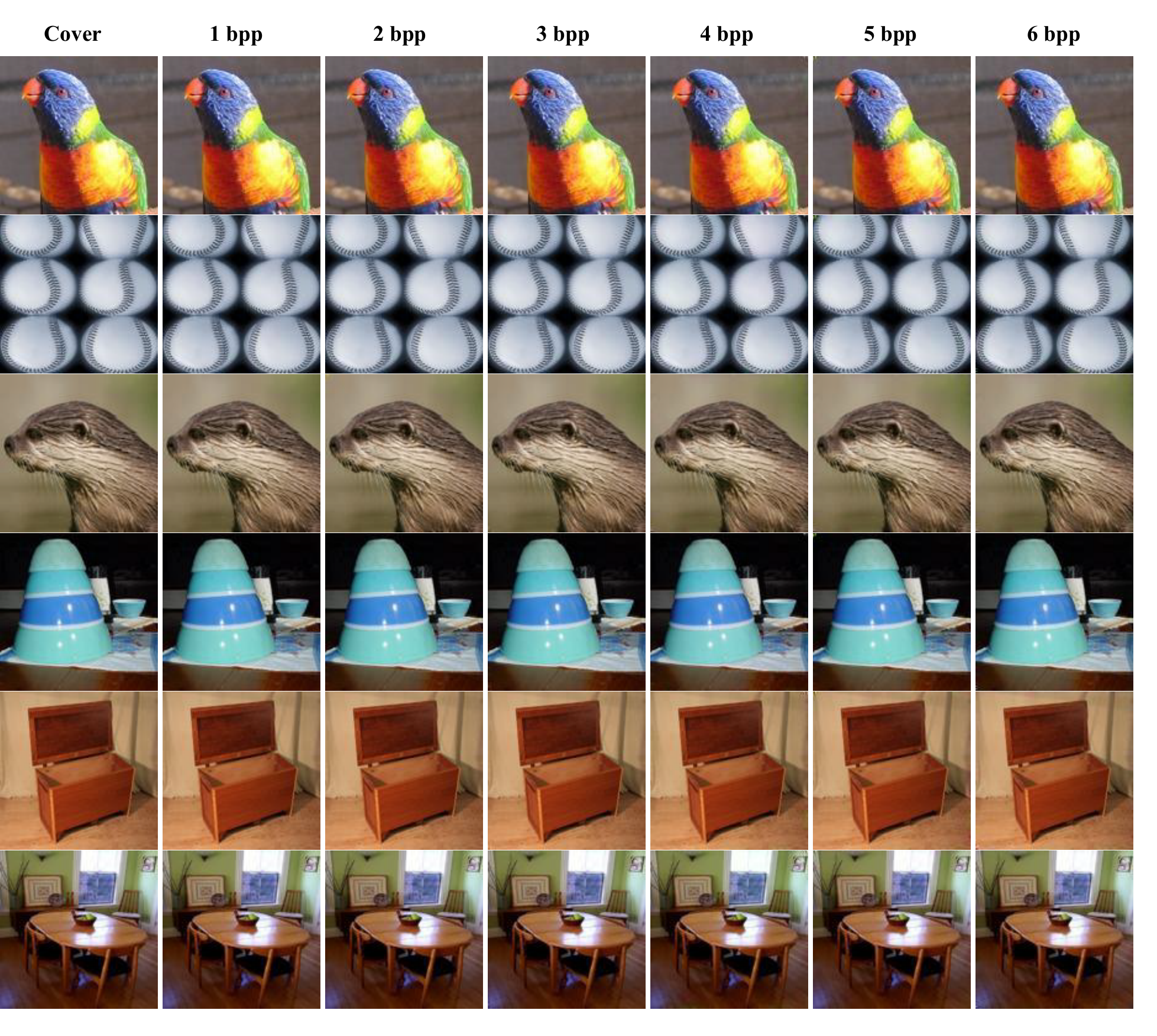}
	\caption{The original image and the stego image generated by CLPSTNet at 1-6 bpp}
	\label{fig3}
\end{figure}

7) Visual test of steganographic images. In order to further validate the quality of steganographic images for the CLPSTNet model, a part of images from ImageNet, ALASKA2, and VOC2012 datasets were selected for testing. This part of the images did not participate in the model training and verified the carrier-confidential images under 1-6 bpp steganographic capacity, and the test results are shown in Figure \ref{fig3}. As can be seen in Figure 11, the steganographic images generated by the CLPSTNet model with 1-6 bpp steganographic capacity are more similar to the original images in terms of color and brightness, which indicates that the model has superior image steganographic quality.

\clearpage

\section{Conclusion}
In this paper, a multi-scale convolutional steganography method, called CLPSTNet, is proposed.Based on GAN, encoding, decoding network, and steganographic analysis network are chosen as the information embedding, information recovery and evaluation parties. The progressive multi-scale convolutional module is used in the encoding network and decoding network, which integrates the idea of curriculum learning algorithm and combines the idea of curriculum learning with the model structure of steganography, which greatly improves the quality of steganographic images and the embedding capacity of images. We use multiple steganography evaluation metrics and steganography analysis network to analyze our proposed model, which realizes the invisibility and undetectability of information. However, the CLPSTNet model also has some weaknesses. First, CLPSTNet can only embed binary information, and cannot be embedded for other data such as images. In addition, the CLPSTNet model does not perform well enough in terms of information recovery accuracy, which appears to decrease with the expansion of information embedding capacity. In the future, we will explore the performance of progressive multi-scale convolutional modules for decoding.


\bibliographystyle{unsrt}  
\bibliography{ref}  


\end{document}